\documentclass[pmlr]{jmlr}

\usepackage{booktabs}
\usepackage[dvipsnames]{xcolor}
\usepackage{tikz}
\usepackage{float}
\usepackage{longtable}

\jmlrvolume{353,}
\jmlryear{2026}
\jmlrworkshop{ACML 2026}
\editors{Andy Song, Bo Han and Sarah Erfani}

\makeatletter
\@ifundefined{@org@Ginclude@graphics}{}{%
  \let\Ginclude@graphics\@org@Ginclude@graphics
}
\makeatother

\title[TallyTrain]%
{TallyTrain: Communication-Efficient Federated Distillation}

\author{%
  \Name{Radhakrishna Achanta} \Email{rachanta@cisco.com}\\
  \Name{Will Reed} \Email{wilreed@cisco.com}\\
  \addr Cisco Systems Inc.
}

\begin{document}

% PMLR assigns the proceedings page range; omit it from the header.
\makeatletter
\let \@jmlrpages \@empty
\makeatother

\maketitle

\begin{abstract}
    Federated learning is bandwidth-bound on two orthogonal axes: model
    size, which limits how often parameter-averaging methods can
    afford to merge, and class count, which makes per-probe
    soft-label distillation prohibitive at large vocabularies. Both
    ceilings tighten as modern systems scale. For $C$ classes, we collapse the class-count axis
    to $\lceil \log_2 C \rceil$ bits per probe by transmitting only each peer's
    $\arg\max$ class index. The resulting protocol, TallyTrain, is not
merely compressed: under non-IID training the voted target stays
peaked, where averaging the softmaxes of under-trained peers flattens
it toward uniform. Across standard benchmarks ($C \in \{10, 100, 2048\}$),
    TallyTrain matches or exceeds soft-label distillation on every
    benchmark we report: $+1.35$~pp over FedMD on CIFAR-100 at $400\times$
    less communication, parity on the CIFAR-10 bridge at equal
    communication, and parity on WikiText-2 at up to $4{,}096\times$ less.
    We also relax the model-size axis by composing the cheap hard-label
    consensus with sparse parameter merges (the bandwidth-bridge variant
    TallyTrain+fa$M$), which improves the accuracy--bandwidth trade-off of
    the FedAvg, FedProx and FedDF baselines, exceeding their accuracy at
    matched communication.
    \end{abstract}


\section{Introduction}

Federated learning (FL) trains a model from data shards held by many
peers without centralizing the raw data. The two dominant paradigms
differ in \emph{what} they communicate.

\textbf{Parameter-space methods} exchange model parameters
(FedAvg~\cite{mcmahan2017communication}) or per-round weight deltas
that play the role of an outer-loop pseudo-gradient
(DiLoCo~\cite{douillard2023diloco}). Both have per-round bandwidth
$\Theta(|W|)$, where $|W|$ is the number of model parameters, which
is impractical for billion-parameter models on edge or mobile
substrates.

\textbf{Function-space methods} (FedMD~\cite{fedmd2019})
exchange per-example predictions on a shared public probe set.
Bandwidth scales as
$\Theta(C\!\cdot\!|\mathcal{D}_{\mathrm{pub}}|)$ per peer per round,
where $C$ is the number of output classes -- independent of model
size, but growing linearly with $C$, which becomes prohibitive for
large-vocabulary tasks (BPE-tokenized language models routinely use
$C \in [2{,}048, 50{,}000]$).

These paradigms share a structural assumption: bandwidth is reduced
by communicating less often, while
the size of each message is treated as fixed (full weights, or a full
$C$-dimensional softmax). This paper proposes to keep communication frequent but make each message tiny: only
the $\arg\max$ class index per probe, one byte for $C\!\le\!256$, two
bytes for $C\!\le\!65{,}536$. Argmax voting keeps the consensus sharp --
the voted histogram is supported on at most $N$ classes, whereas
averaging the softmaxes of under-trained peers flattens the target
toward uniform and turns distillation into an entropy-increasing
pull. We call the resulting protocol
\textbf{TallyTrain} -- each round, peers \emph{tally} their argmax
votes on the public probe set into a single consensus histogram and then
\emph{train} against it -- and show that, combined with sparse parameter
merges, it advances the bandwidth--accuracy Pareto frontier of
federated learning across two modalities and three class counts. Our contributions are as follows:

% \paragraph{Contributions.}
\noindent\textbf{(1) A hard-label communication primitive that matches
or beats soft labels.} We introduce TallyTrain, which treats the voting
histogram of $N$ peers' argmax predictions over a public probe set as
the consensus distribution for distillation, and show the $\arg\max$
channel carries essentially all the signal soft labels do -- beating
soft-label distillation on CIFAR-100 and reaching parity elsewhere, at a per-probe
bandwidth ratio that grows linearly in $C$
(\S\ref{sec:t1}, \S\ref{sec:scaling}, \S\ref{sec:charlm}).

\noindent\textbf{(2) Two operational regimes.}
The same primitive runs in two modes. In the \emph{pure} mode,
collaboration happens entirely through the hard-label channel -- the
lowest-bandwidth option, whether or not the public probes carry
ground-truth labels (\S\ref{sec:t1}, \S\ref{sec:secondary}). In the
\emph{bandwidth-bridge} mode, TallyTrain interleaves occasional
FedAvg parameter merges with the hard-label channel to reach
parameter-averaging accuracy at a fraction of its communication
(\S\ref{sec:t3}, \S\ref{sec:charlm}); at a merge frequency comparable
to standard FedAvg it improves tail accuracy by $+17$~pp at equal
bandwidth.

\noindent\textbf{(3) A contractive theory of hard-label distillation:}
a function-space contraction lemma, a Condorcet bound on the voted
top-1, and a sharpness separation between voted and averaged targets
(\S\ref{sec:method}).

Note that our target is bandwidth, not compute: each peer
holds a full model and runs independent local gradient descent, so total compute
scales linearly in $N$. Topology and scale-up ($N\!\gg\!10$, sparse
gossip) are discussed in \S\ref{sec:future}.

\section{Related Work}
\label{sec:related_work}

Distributed and federated learning methods \cite{kairouz2019advances} differ not only in topology (centralized vs.\ peer-to-peer) but in the \emph{object of communication}. Classical decentralized federated learning (DFL) exchanges parameters, gradients, or compressed model deltas; distillation-based methods instead communicate \emph{prediction-space objects} -- logits, softened targets, prototypes, centroids, or function values -- decoupling collaboration from exact parameter alignment. This distinction matters most when clients differ in architecture, capacity, data distribution, or communication budget. We review the relevant literature through this lens, beginning with the distillation lineage and then positioning it against optimization-centric DFL.

\subsection{Knowledge Distillation as a Collaboration Primitive}

Knowledge distillation (KD) lets a student learn from a teacher's softened predictions \cite{hinton2015distilling}; its implication for distributed learning is that collaboration can happen in \emph{function space} -- the shared object is a predictive distribution, not a weight vector -- which accommodates heterogeneous models and can cut communication when only compact predictive summaries are exchanged. \emph{Federated distillation} (FD) \cite{jeong2018communication} exchanges averaged per-class logits over the federation. \emph{FedMD} \cite{fedmd2019} formalises the public-probe-set version: clients hold private data and possibly different architectures, share a small public reference set, and use the aggregated logits on it as distillation targets. \emph{FedDF} \cite{lin2020ensemble} instead ensembles server-side -- clients communicate full models and a server-side student is distilled from their soft predictions on a public set, so its per-round cost is $\Theta(|W|)$, not a per-probe vector. \emph{Cronus} \cite{chang2019cronus} shares length-$C$ \emph{soft} labels on a public set, aggregated \emph{robustly} (e.g.\ coordinate-wise median) for Byzantine resilience and privacy. \emph{FedKD} \cite{fedkd2021} keeps the KD viewpoint but communicates a small shared student model with SVD-compressed gradients -- again a model-space payload. These methods are server-coordinated; together they show distillation can bridge heterogeneity and cut communication, but none gives a fully decentralized protocol whose per-probe payload scales with $\log_2 C$ rather than $C$.

\subsection{Decentralized Distillation for Heterogeneous Learning}

Several lines transplant the FedMD intuition into peer-to-peer settings.
\emph{DFML} \cite{dfml2024} is serverless and avoids public data: neighbours distill from each other on a cyclic supervised/distillation schedule, at the cost of sensitivity to neighbourhood quality.
\emph{DeSA} \cite{desa2024} builds a shared alignment substrate from \emph{synthetic} anchors, avoiding real public data but paying a synthesis and anchor-quality cost.
\emph{DFedCAD} \cite{dfedcad2025} replaces full predictions with class-centroid summaries -- cheaper, but coarser in the uncertainty transmitted.
The design space thus decomposes by \emph{what is shared} -- public anchors, peer predictions, synthetic anchors, or centroids -- each with different assumptions about auxiliary data, heterogeneity, and communication.

\subsection{Optimization-Centric Decentralized Federated Learning}

In parallel, a strong line of work improves DFL in parameter space. DSpodFL \cite{dspodfl2025}, DPFL \cite{dpfl2025}, NTK-DFL \cite{ntkdfl2025}, FedSPD \cite{fedspd2025}, GFlat \cite{gflat2026}, PaME \cite{pame2026}, and Spod-GT \cite{spodgt2026} contribute sporadic-communication convergence, learned collaboration graphs, NTK stabilization, soft-cluster personalization, flat-minima generalization, partial-message bandwidth reduction, and gradient tracking on directed graphs. They are preferable when clients can share a common model family and the challenge is systems efficiency, but their parameter-space messages presuppose that the transmitted coordinates mean comparable things across clients, limiting cross-architecture transfer. A complementary strand widens DFL's scope beyond accuracy -- privacy-utility trade-offs \cite{fdpdfl2025}, strategic manipulation \cite{compadvdfl2025}, and satellite-network topology \cite{orbitdfl2026}. More broadly, decentralized learning spans sequential-decision and estimation under the same constraints: peer-to-peer linear bandits \citep{korda2016distributed,mahadik2020fast} and online quantification \citep{narasimhan2016stochastic}.

\paragraph{Positioning TallyTrain.}
TallyTrain targets the engineering bottlenecks left open above. Unlike
methods that communicate model-sized objects -- FedAvg and DiLoCo
(parameters or pseudo-gradients) and FedDF (server-side ensemble
distillation over client models) -- it operates entirely in function
space, at $\Theta(\log C)$ per probe independent of $|W|$, sidestepping
the weight-merging barrier on non-IID data. Against per-probe
soft-label distillation (FedMD, Cronus) it differs on three counts: it transmits a single $1$--$2$-byte $\arg\max$ index
($\lceil\log_2 C\rceil$ bits) rather than a $4C$-byte softmax -- a
$40\times$--$4{,}096\times$ per-probe reduction for $C\!\in\![10,2048]$;
it aggregates by plurality vote of indices rather than a (robust) mean
of soft vectors; and it composes with sparse parameter merges in the
bandwidth-bridge variant (\S\ref{sec:t3}). In the non-Byzantine regime
a robust mean reduces to the ordinary FedMD average -- a
median-aggregation ablation ties plain FedMD (CIFAR-100 tail $31.95$ vs
$31.81$; Table~\ref{tab:t1}) -- so Cronus's empirical stand-in here is
our FedMD baseline
(Tables~\ref{tab:t1}--\ref{tab:lm}). Unlike DFML and other peer-to-peer
mutual-distillation protocols, the inter-peer payload is \emph{only}
the hard prediction; no soft signals, anchors, or parameters cross the
wire. And unlike the decentralized bandit and quantification methods
above, TallyTrain performs federated \emph{supervised} training --
every peer minimizes the same supervised loss on a fixed private shard
-- rather than balancing exploration against exploitation or estimating
a population prevalence. The protocol is therefore most attractive
whenever $|W|$ or $C$ is large enough that float-vector or parameter
exchange dominates the communication budget (\S\ref{sec:charlm},
\S\ref{sec:scaling}).

\section{Method}
\label{sec:method}

% Paradigm comparison table (replaces the earlier four-panel schematic):
% methods in columns, communication attributes in rows, so the payload
% differences are explicit in the table itself rather than caption-only.
\begin{table}[t]
\centering
\small
\resizebox{\textwidth}{!}{%
\begin{tabular}{l cccc}
\toprule
 & FedAvg & DiLoCo & FedMD & \textbf{TallyTrain} \\
\midrule
Communicated object          & weights $W$ & pseudo-gradient $\Delta W$ & softmax ($\in\mathbb{R}^C$) & $\arg\max$ index \\
Communication space          & parameter   & parameter                 & function                   & function \\
Message size                 & $4|W|$ B    & $4|W|$ B                  & $4C$ B/probe               & $1$--$2$ B/probe \\
Grows with model size $|W|$? & yes         & yes                       & no                         & no \\
Grows with class count $C$?  & no          & no                        & $\propto C$                & $\propto\lceil\log_2 C\rceil$ \\
\bottomrule
\end{tabular}%
}
\caption{Federated learning paradigms by \emph{what} they communicate.
FedAvg and DiLoCo exchange parameter-space objects whose size grows
with the model ($\Theta(|W|)$); FedMD and TallyTrain exchange
function-space objects whose size grows with the class count $C$.
TallyTrain sends only the $\arg\max$ class index ($1$--$2$ bytes, or
$\lceil\log_2 C\rceil$ bits), collapsing the $C$ axis: at the
WikiText-2 vocabulary $C\!=\!2048$ this is $4096\times$ smaller per
probe than FedMD's softmax. Topology is orthogonal to this choice --
any of these protocols runs on full-mesh, ring, or gossip without
algorithmic change (\S\ref{sec:method}).}
\label{tab:paradigms}
\end{table}

TallyTrain is built around a single design choice -- argmax voting over a
shared public probe set -- with two operating variants that compose
with it: a labelled-public-set hybrid that anchors peers to ground
truth, and a bandwidth-bridge variant that interleaves periodic
FedAvg parameter merges. Table~\ref{tab:paradigms} situates the
primitive against existing federated learning paradigms by what each
communicates and how that payload scales.

\subsection{Function-space alignment}
\label{sec:functionspace}

Let $\mathcal{D}_n$ denote peer $n$'s private non-IID shard and
$\mathcal{D}_{\mathrm{pub}}$ a public probe set shared by all peers,
with each peer holding a predictor $f_n(\cdot;W_n)$ from the input
space $\mathcal{X}$ to the probability simplex $\Delta^{C-1}$.
Weight-averaging $W_g\!=\!\tfrac{1}{N}\sum_n W_n$ inherits the
loss-surface mismatch endemic to non-linear models on non-IID data.
TallyTrain aligns the \emph{functions} on $\mathcal{D}_{\mathrm{pub}}$
instead: two models that agree on all
$x\!\in\!\mathcal{D}_{\mathrm{pub}}$ implement the same predictor on
any test distribution dominated by it.

\subsection{The communication primitive: argmax voting}
\label{sec:primitive}

For each $x \in \mathcal{D}_{\mathrm{pub}}$, peer $n$ broadcasts only the
top-1 prediction
\begin{equation*}
    y_n(x) = \arg\max f_n(x; W_n) \in \{1,\dots,C\}.
\end{equation*}
A class index requires $\lceil\log_2 C\rceil$ bits, transmitted
byte-aligned ($b_C\!=\!1$ B for $C\!\le\!256$, $2$ B for
$C\!\le\!65{,}536$). Against $4C$-byte 32-bit soft labels the
implementation ratio is $\rho_C\!=\!4C/b_C$
($\rho_{10}\!=\!40,\rho_{100}\!=\!400,\rho_{2048}\!=\!4096$); the
bit-packed lower bound is
$\rho_C^{\star}\!=\!32C/\lceil\log_2 C\rceil$. The saving grows
linearly in $C$.

Across peers, the hard-label predictions form the empirical voting
histogram
\begin{equation*}
    \bar H(x) = \frac{1}{N}\sum_{n=1}^N e_{y_n(x)} \in \Delta^{C-1},
\end{equation*}
where $e_c$ is the $c$-th basis vector. $\bar H(x)$ is a valid
probability distribution and serves as the consensus target for
distillation.

\paragraph{Decentralized $\equiv$ centralized in trajectory.} Under
full-mesh, each peer locally averages $\{y_m(x)\}$ to obtain the same
$\bar H(x)$ a centralized relay would compute, so the per-peer
distillation gradient is identical and the two topologies share their
optimization trajectory; they differ only in the wire pattern
(all-to-all vs.\ star). All numbers in \S\ref{sec:experiments} are
therefore simultaneously valid for both interpretations. Per-peer
per-probe bandwidth scales as $(N\!-\!1)b_C$ bytes for decentralized
hard labels vs.\ $4C$ bytes for a centralized soft-label star, so hard
labels are cheaper whenever $N\!\le\!4C/b_C\!+\!1$ (e.g., $N\!\le\!41$
for $C\!=\!10$, $N\!\le\!401$ for $C\!=\!100$); beyond that,
sparse-gossip topologies are required (\S\ref{sec:future}).

\subsection{Theoretical analysis}
\label{sec:theory}

Three results explain TallyTrain's behavior; we state each with a proof
sketch. Throughout, let
$\bar H^{(r)}(x) = \frac{1}{N}\sum_n e_{\arg\max f_n^{(r)}(x)}$ be the
voting histogram and
$\overline{D_{\mathrm{KL}}}^{(r)}$ the mean pairwise function-space
disagreement
$\frac{1}{N(N-1)}\sum_{m\neq n}\mathbb{E}_x\,
D_{\mathrm{KL}}(f_n^{(r)}\|f_m^{(r)})$.

\paragraph{Lemma 1 (Function-space contraction).} Under standard
$L$-smoothness, bounded SGD variance and non-degenerate predictors,
with $\eta L\!\le\!1/2$,
\begin{equation*}
\mathbb{E}\!\left[\,\overline{D_{\mathrm{KL}}}^{(r+1)}\,\right]
\le \big(1 - (1-\alpha)\,\eta\,\mu_{\mathrm{KL}}\big)\,\overline{D_{\mathrm{KL}}}^{(r)}
+ \tfrac{\eta^2 \sigma_g^2}{\beta_0},
\end{equation*}
so peers contract toward agreement on $\mathcal{D}_{\mathrm{pub}}$;
predictors with $\overline{D_{\mathrm{KL}}}\!\to\!0$ are
indistinguishable on any test distribution dominated by it.
(Lemma~1 bounds mutual agreement, not correctness; Proposition~1 below
supplies the latter, which is why the unlabelled A.2 mode can still
collapse.)
\emph{Proof sketch.} The KL gradient in the predictor is bounded below
by the strong-convexity constant $\mu_{\mathrm{KL}}$ of KL on the $\beta_0$-interior
of the simplex; one distillation step moves each peer along the
negative KL gradient with step size $(1-\alpha)\eta$, giving the
contraction factor $1-(1-\alpha)\eta\mu_{\mathrm{KL}}$, while the
$\eta^2\sigma_g^2/\beta_0$ term follows from the bounded-variance
assumption and the coordinate lower bound $\beta_0$. The argument
adapts standard stochastic-approximation contraction reasoning (as in
local-SGD analyses~\citep{stich2018local,khaled2020tighter}) to the
function-space KL-distillation operator.

\paragraph{Proposition 1 (Voting accuracy; Condorcet).} If each peer's
top-1 accuracy exceeds $\bar p > 1/2$ and the
$\{\arg\max f_n(x)\}$ are independent (not necessarily identically
distributed), the majority vote
hits the true class $c$ with probability
$\ge 1 - \exp(-2N(\bar p-1/2)^2)$. Here $\bar p\!>\!1/2$ is the
binary-majority special case: plurality requires only that $c$ be
\emph{modal}, and with the residual $1\!-\!\bar p$ spread over
$C\!-\!1$ alternatives this holds well below $1/2$ once
$C\!\gg\!2$ -- the regime of our CIFAR-100 and WikiText-2 cells.
\emph{Proof.} Hoeffding's inequality bounds the probability that fewer
than $N/2$ of $N$ independent Bernoulli($p_n\!\ge\!\bar p$) trials
succeed by $\exp(-2N(\bar p-1/2)^2)$; the event ``mode correct''
contains ``majority correct'', so the same bound applies.

\paragraph{Proposition 2 (Consensus sharpness).} With
$\bar P(x)\!=\!\frac{1}{N}\sum_n f_n(x)$ the soft consensus and
$\mathcal{H}(\cdot)$ Shannon entropy in nats,
\begin{equation*}
\mathcal{H}(\bar H) \le \log\min(N,C)
\quad\text{while}\quad
\mathcal{H}(\bar P) \ge \tfrac{1}{N}\textstyle\sum_n \mathcal{H}(f_n),
\end{equation*}
so once mean peer entropy exceeds $\log\min(N,C)$ -- only $2.30$ nats
at $N\!=\!10$, against $\log C\!=\!4.61$ at $C\!=\!100$ -- the voted
target is \emph{strictly} sharper. Under-trained peers therefore
flatten $\bar P$ toward uniform while $\bar H$ stays peaked, so
distilling to $\bar P$ adds an entropy-increasing pull that $\bar H$
avoids. \emph{Proof.} $\bar H$ has mass on at most $\min(N,C)$ distinct
indices and a distribution on $k$ atoms has entropy $\le\log k$; the
second bound is concavity of $\mathcal{H}$. No distributional
assumptions are needed.

\subsection{Algorithm and operating modes}
\label{sec:algorithm}

Algorithm~\ref{alg:dhlfd} is a single per-peer procedure with two
orthogonal operating modes that are selected by setting the KL weight
$\alpha$ (with optional KL-decay) and the FedAvg cadence $M$.

\begin{algorithm2e}[t]
\footnotesize
\DontPrintSemicolon
\caption{TallyTrain (per-peer, round $r$)}
\label{alg:dhlfd}
\KwIn{private data $\mathcal{D}_n$, public probes
$\mathcal{D}_{\mathrm{pub}}$ (labels $y_{\mathrm{pub}}$ if available), rounds $R$, local steps $K$, sample size
$B_{\mathrm{sample}}$, warm-up $W_{\mathrm{up}}$, KL weight $\alpha$, optional KL
decay window $T$, optional FedAvg cadence $M$}
Initialize local weights $W_n$\;
\For{round $r = 1, \dots, R$}{
  \For{$k = 1, \dots, K$}{
    Sample $(x, y) \sim \mathcal{D}_n$\tcp*{local SGD on private data}
    $W_n \leftarrow W_n - \eta\,\nabla \mathcal{L}_{\mathrm{CE}}(f(x; W_n), y)$\;
  }
  \If{$r > W_{\mathrm{up}}$}{
    Sample $\mathcal{S}_r \subset \mathcal{D}_{\mathrm{pub}}$, $|\mathcal{S}_r|\!=\!B_{\mathrm{sample}}$\tcp*{consensus channel}
    Broadcast $Y_n = \arg\max f(\mathcal{S}_r; W_n)$; receive $\{Y_m\}_{m\neq n}$\tcp*{$b_C$ bytes/probe}
    $\bar H \leftarrow \tfrac{1}{N}\sum_{m=1}^{N} e_{Y_m}$ on $\mathcal{S}_r$\tcp*{voting histogram}
    $\lambda_r \leftarrow \mathrm{KL\text{-}decay}(r; W_{\mathrm{up}}, T)$\tcp*{$\equiv 1$ if $T$ unset}
    $\mathcal{L} = \alpha\,\mathcal{L}_{\mathrm{CE}}(f(\mathcal{S}_r; W_n), y_{\mathrm{pub}}) + \lambda_r(1\!-\!\alpha)\,D_{\mathrm{KL}}(\bar H \| f(\mathcal{S}_r; W_n))$\;
    $W_n \leftarrow W_n - \eta\,\nabla \mathcal{L}$\;
  }
  \If{$M > 0$ \textnormal{and} $r \bmod M = 0$}{
    $W_n \leftarrow \tfrac{1}{N}\sum_{m=1}^N W_m$\tcp*{Mode B: FedAvg merge}
  }
}
\KwRet $W_n$
\end{algorithm2e}

\paragraph{Mode A -- pure function-space distillation ($M\!=\!0$).}
The protocol runs entirely on the hard-label channel. Two sub-modes
cover the public-set assumption.

\noindent\emph{A.1 Labelled probes ($\alpha\!>\!0$, no decay).} When
$\mathcal{D}_{\mathrm{pub}}$ carries ground-truth labels, the CE term
keeps each peer's top-1 class modal on the probes, so the
contraction of Lemma~1 converges toward the labelled
signal rather than to an arbitrary agreement point. Empirically $\alpha\!=\!0.5$ transfers across
CIFAR-10/100 without retuning (Tables~\ref{tab:t1}~\&~\ref{tab:t3},
top TallyTrain row).

\noindent\emph{A.2 Unlabelled probes ($\alpha\!=\!0$, KL-decay).} When
$\mathcal{D}_{\mathrm{pub}}$ has no useful labels (e.g.\ drawn from a
different distribution), the CE anchor is absent and naïve pure-KL
training drifts into Condorcet-class collapse. A linear decay
$\lambda_r\!=\!\max(0,1\!-\!(r\!-\!W_{\mathrm{up}})/T)$ for $r\!>\!W_{\mathrm{up}}$, with
$\lambda_r\!=\!0$ thereafter, halts the drift while still letting the
consensus channel transfer useful knowledge; the choice of $T$ is
robust across an order of magnitude (\S\ref{sec:secondary}).

\paragraph{Mode B -- function-space stabilizer for parameter-space
averaging ($M\!>\!0$).} Pure function-space distillation plateaus
below FedAvg's accuracy ceiling: hard-label voting cannot transfer
the rich feature representations parameter averaging does. Setting
$M\!>\!0$ interleaves a FedAvg merge every $M$ rounds, lifting peers
toward the parameter-space ceiling while the hard-label channel keeps
them tightly synchronized between merges. The combination has
substantially lower cross-peer drift than FedAvg at the same $M$ and
reaches strictly higher tail accuracy at the same bandwidth. We write
\emph{fa$M$} for ``FedAvg every $M$ rounds'' and use it for both the
bridge variant TallyTrain+fa$M$ and the parameter-only baseline
FedAvg-fa$M$ (Tables~\ref{tab:t3}~\&~\ref{tab:lm};
Pareto frontier in Figure~\ref{fig:pareto}). It is convenient to read
the merge interval $M$ in the units of
\citet{mcmahan2017communication}'s FedAvg -- $E$, the number of local
epochs each peer trains between merges. On our CIFAR-10 non-IID setup
one merge every $M$ rounds is $E\!\approx\!M/30$ local epochs, so
$\mathrm{fa}\!=\!200$ ($E\!\approx\!6.7$) is the closest cell to the
McMahan CIFAR default of $E\!=\!5$ (fa $\!\approx\!150$), while
$\mathrm{fa}\!=\!500$ ($E\!\approx\!16.7$) merges $2.5\times$ less often.

Any A-mode composes with any $M$. Most
deployments fall into one of three combinations -- A.1 with $M\!=\!0$
(\S\ref{sec:t1}), A.2 with $M\!=\!0$ (\S\ref{sec:secondary}), and
A.1 with $M\!>\!0$ (\S\ref{sec:t3}, \S\ref{sec:charlm}); the latter is
our recommended deployment when absolute accuracy matters more than
the lowest possible bandwidth.

\subsection{Design rules}
\label{sec:design-rules}

Five hyperparameters drive deployment, with the following empirical
defaults that we use throughout. The distillation step operates on a
stochastic subset of size $B_{\mathrm{sample}}$, so per-peer per-round
egress is $(N\!-\!1)B_{\mathrm{sample}}b_C$ bytes regardless of
$|\mathcal{D}_{\mathrm{pub}}|$; the consensus channel is disabled for
the first $W_{\mathrm{up}}$ warm-up rounds so peers become individually
informative before influencing each other ($W_{\mathrm{up}}\!=\!300$ in-distribution,
$W_{\mathrm{up}}\!=\!1500$ cross-distribution). Local steps $K\!=\!5$: weight-exchange
methods are forced to large $K$ to amortize $\Theta(|W|)$ sync cost,
but TallyTrain removes that constraint
\citep{stich2018local,khaled2020tighter}. Public set $|\mathcal{D}_{\mathrm{pub}}|\!=\!2{,}000$ for
CIFAR-10 and $10{,}000$ for CIFAR-100, sized so $\mathcal{D}_{\mathrm{pub}}$
$\epsilon$-covers the test manifold. Sub-sample $B_{\mathrm{sample}}$
is best read as a regularization knob: empirically the per-class
coverage rate $\kappa\!=\!B_{\mathrm{sample}}/C \in [1,2]$ is optimal
(our CIFAR-100 headline fixes $B_{\mathrm{sample}}\!=\!32$, i.e.\
$\kappa\!=\!0.32$, for cross-task comparability); see
\S\ref{sec:secondary}. Number of peers
$N$: Proposition~1 saturates at $N\!\approx\!1/(2(\bar p\!-\!1/2)^2)$
($\approx\!50$ for $\bar p\!\approx\!0.6$); the hard-label primitive
itself remains cheaper than soft labels up to the crossover
$N\!\le\!4C/b_C\!+\!1$ derived in \S\ref{sec:primitive}.

\section{Experiments}
\label{sec:experiments}

We evaluate TallyTrain against soft-label distillation
(FedMD~\citep{fedmd2019}), centralized ensemble distillation
(FedDF~\citep{lin2020ensemble}), parameter-averaging baselines
(FedAvg~\citep{mcmahan2017communication},
FedProx~\citep{li2018federated}), and a no-communication local-only
baseline on CIFAR-10/100 with Dirichlet non-IID splits
($\alpha_{\mathrm{Dir}}\!=\!0.5$; $N\!=\!10$ peers, ResNet-18, AdamW
$\eta\!=\!10^{-3}$ with weight decay $5\!\cdot\!10^{-4}$, local batch
$B\!=\!32$, $K\!=\!5$, $R\!=\!3000$). Each multi-seed cell uses seeds
$\{0,1,2\}$; we report the tail mean $\pm$ std across these seeds, and
the peak (best round-level mean), over the last 100 logged rounds
(CIFAR) or 600 rounds (WikiText-2, three full fa$=200$ merge cycles). Peers use distinct per-peer init RNG so voting has a
non-degenerate signal at round zero. ResNet-18 BN statistics are
stripped for \texttt{vmap} batching (a $<\!1$~pp uniform cost, so
relative comparisons are unaffected); the WikiText-2 TinyGPT uses
LayerNorm and is unaffected. Bandwidth is cumulative per-peer bytes on
the wire: $(N\!-\!1)b$ per probe for function-space messages
($b\!=\!b_C$ hard, $4C$ soft), and $2\cdot 4|W|$ per parameter merge
(one upload, one download, since averaging is a reduction), at the
protocol cadence of $R/M$ merges. This prices our own channel at
full-mesh cost while granting each parameter-space baseline its
cheapest topology.

\subsection{Hard labels match or beat soft labels at $400\times$ less bandwidth}
\label{sec:t1}

Table~\ref{tab:t1} reports the labelled-hybrid cell on CIFAR-100
non-IID ($\alpha\!=\!0.5$, $W_{\mathrm{up}}\!=\!300$, $B_{\mathrm{sample}}\!=\!32$,
$|\mathcal{D}_{\mathrm{pub}}|\!=\!10{,}000$). TallyTrain and FedMD share
the labelled-hybrid loss; the $\alpha\!=\!1$ row is the public-CE-only
baseline that isolates each method's consensus contribution.

\begin{table}%[H]
\centering
\small
\resizebox{\textwidth}{!}{%
\begin{tabular}{l l |cccc|c}
\toprule
Method & & Tail acc.\ (\%) & Peak acc.\ (\%) & Bytes/probe & Payload (MB) & Reduction \\
\midrule
\textbf{TallyTrain} & ($\alpha=0.5$) & $\mathbf{33.16 \pm 0.27}$ & $33.44 \pm 0.24$ & 1 & 0.78 & $\mathbf{400\times}$ \\
FedMD &($\alpha=0.5$) & $31.81 \pm 0.10$ & $32.16 \pm 0.13$ & 400 & 311.0 & 1$\times$ \\
FedMD, median agg.\ & ($\alpha=0.5$) & $31.95 \pm 0.05$ & $32.64 \pm 0.07$ & 400 & 311.0 & 1$\times$ \\
No consensus & ($\alpha=1.0$) & $32.03 \pm 0.06$ & $32.48 \pm 0.09$ & 0 & 0 & --- \\
\bottomrule
\end{tabular}%
}
\caption{CIFAR-100 non-IID hybrid, 3 seeds each. The $\alpha\!=\!1.0$
row is the public-CE-only baseline (consensus weight zero, so identical
regardless of how consensus is computed); ``median agg.'' swaps FedMD's
mean for a coordinate-wise median (the Cronus-style robust aggregator)
and leaves it unchanged ($31.95$ vs $31.81$), i.e.\ Cronus reduces to
FedMD without Byzantine peers. Findings (i)--(iii) in text.}
\label{tab:t1}
\end{table}

\noindent We make three findings: (i) TallyTrain beats FedMD by $1.35$~pp at $400\times$
less bandwidth -- about $5\times$ the per-method seed std, with
TallyTrain's tail above FedMD's peak. (ii) \emph{Soft-label consensus
contributes nothing on CIFAR-100}: the $\alpha\!=\!1$ no-consensus
baseline ($32.03\%$) sits $0.22$~pp \emph{above} FedMD; the soft-label
average of $10$ under-trained $100$-class softmaxes flattens toward
uniform (Proposition~2), so distilling to it injects entropy, while
the voted target stays peaked and recovers $+1.13$~pp.
(iii) Richer mechanisms add nothing over uniform argmax voting:
disagreement-weighted distillation and asymmetric peer weighting reach
tail $32.74$ and $32.85\%$ against uniform voting's $32.94\%$ (1 seed),
so uniform voting is already at the function-space ceiling.

\subsection{Bandwidth-bridge variant: distillation as peer stabilizer}
\label{sec:t3}

Table \ref{tab:t3} reports CIFAR-10 non-IID under the bandwidth-bridge
variant, which interleaves a FedAvg parameter merge every $M$ rounds on
top of the distillation channel. We compare TallyTrain and FedMD (with
$M=200$ merges) against pure FedAvg at four merge frequencies.

\begin{table}%[H]
\centering
\small
\resizebox{\textwidth}{!}{%
\begin{tabular}{l|cccc|c}
\toprule
Method & $M$ & Tail acc.\ (\%) & Peak acc.\ (\%) & Cross-peer std & Payload (MB) \\
\midrule
Local-only         & --       & $47.79 \pm 1.68$ & $48.48 \pm 1.62$ & 7.16   & 0     \\
\midrule
FedDF              & 1   & $52.34 \pm 4.79$ & $66.02 \pm 1.62$ & 0$^\dagger$ & 268{,}359 \\
\midrule
FedAvg-fa50        & 50   & $60.51 \pm 1.52$ & $79.60 \pm 0.55$ & 5.56 & 5{,}367 \\
FedAvg-fa200       & 200  & $54.63 \pm 1.77$ & $77.04 \pm 0.99$ & 6.28 & 1{,}342 \\
FedAvg-fa500$^\ddagger$ & 500  & $50.69 \pm 1.85$ & $74.04 \pm 1.22$ & 6.50 & \phantom{0,}537 \\
\midrule
FedProx-fa50       & 50   & $58.79 \pm 2.14$ & $78.69 \pm 1.19$ & 6.02 & 5{,}367 \\
FedProx-fa200      & 200  & $51.71 \pm 2.00$ & $74.91 \pm 1.20$ & 6.49 & 1{,}342 \\
\midrule
FedAvg-fa200 + public CE & 200 & $70.92 \pm 0.76$ & $76.68 \pm 0.55$ & 1.18 & 1{,}342 \\
\midrule
FedMD+fa200        & 200 & $72.68 \pm 0.69$ & $78.31 \pm 0.72$ & 0.93 & 1{,}357 \\
TallyTrain+fa200 & 200 & $71.92 \pm 0.38$ & $77.74 \pm 0.45$ & $1.00$ & 1{,}343 \\
\bottomrule
\end{tabular}%
}
\caption{CIFAR-10 non-IID, $N\!=\!10$, 3 seeds. Cross-peer std is the
mean within-run std of peer test accuracies over the last 100 rounds.
FedProx uses $\mu\!=\!0.001$, the best of $\{0.001,0.01,0.1\}$;
FedDF~\citep{lin2020ensemble}. $^\dagger$FedDF's single server-side
student makes cross-peer std formally zero.
$^\ddagger$fa$500\!\leftrightarrow\!E\!\approx\!16.7$ local
epochs/merge; the McMahan default ($E\!=\!5$) is nearest our fa$200$
cell. The fa$200$ rows communicate essentially the same bytes,
and the `public~CE' row adds ground-truth public-set loss at no extra
communication (see \S\ref{sec:t3}). TallyTrain+fa$M$ beats every
FedAvg/FedProx/FedDF cell with the smallest seed std ($\sigma\!=\!0.38$).}
\label{tab:t3}
\end{table}

\paragraph{What the bridge does, and does not, show.} The $+17$~pp gap
from FedAvg-fa200 ($54.6\%$) to TallyTrain+fa200 ($71.9\%$) is
\emph{not} a hard-vs-soft result. Most of it -- $\sim\!16$~pp -- comes
simply from peers also training on the public set: the
`FedAvg-fa200 + public~CE' row reaches $70.9\%$ with ground-truth
supervision and no consensus channel. The consensus channel adds only
the last $\sim\!1$~pp, where hard-label voting ($71.9\%$) and
soft-label distillation ($72.7\%$) are tied. The bridge is therefore a
\emph{composition} result: interleaving the cheap
$\lceil\log_2 C\rceil$-bit channel with sparse merges reaches
parameter-averaging accuracy at parameter-averaging bandwidth, with far
lower cross-peer drift ($1.00$ vs $6.28$). The claim that voting
\emph{beats} soft labels lives in the pure regime
(Table~\ref{tab:t1}: $+1.13$ vs $-0.22$~pp).

\begin{figure}[!htbp]
\centering
\includegraphics[width=0.75\linewidth]{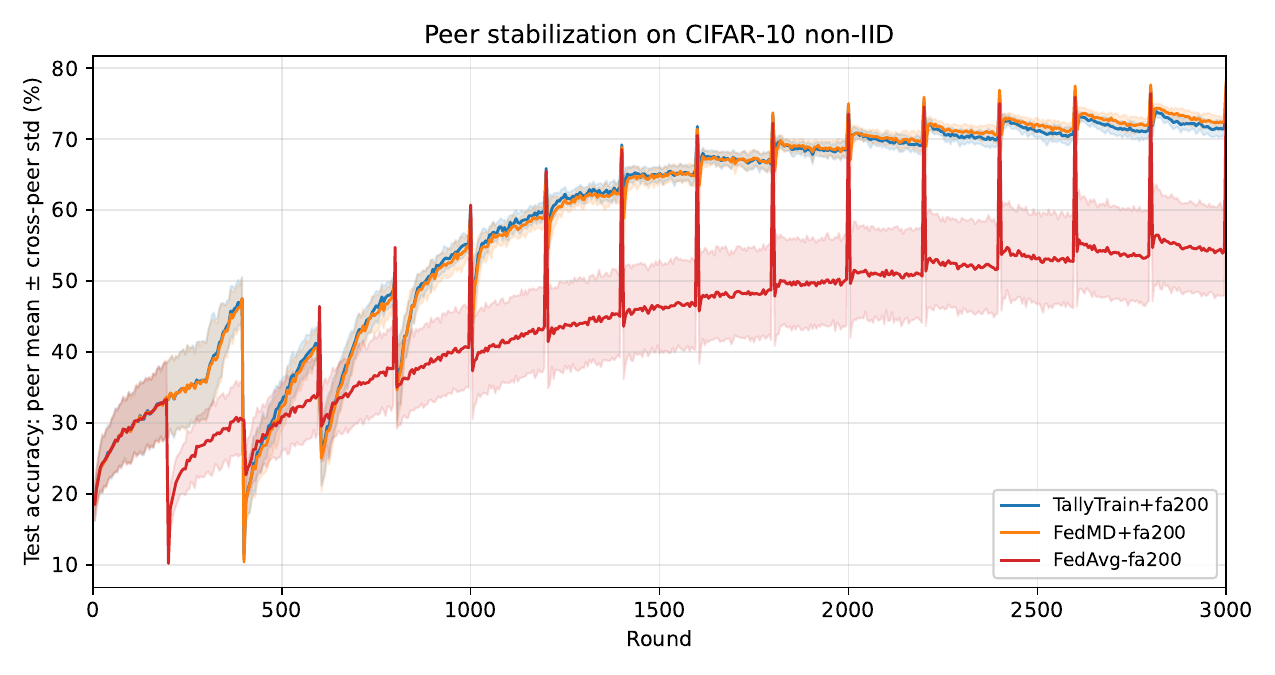}
\caption{Peer stabilization on CIFAR-10 non-IID, 3 seeds. Lines are
peer means, bands are $\pm$1 cross-peer std. TallyTrain+fa$M$ and
FedMD+fa200 keep thin bands throughout; FedAvg-fa200 at the same merge
cadence oscillates between merges. The cheap hard-label channel
stabilizes peers between merges at $\lceil\log_2 C\rceil$ bits per
probe (soft-label distillation does so equally well).}
\label{fig:peer_stabilizer}
\end{figure}

Figure \ref{fig:peer_stabilizer} visualizes this: TallyTrain+fa200 and
FedMD+fa200 ride together along the top with thin peer bands, while
FedAvg-only sits lower with much wider bands and the
``spike-and-recover'' inter-merge drift that distillation prevents.

\begin{figure}[!htbp]
\centering
\includegraphics[width=0.75\linewidth]{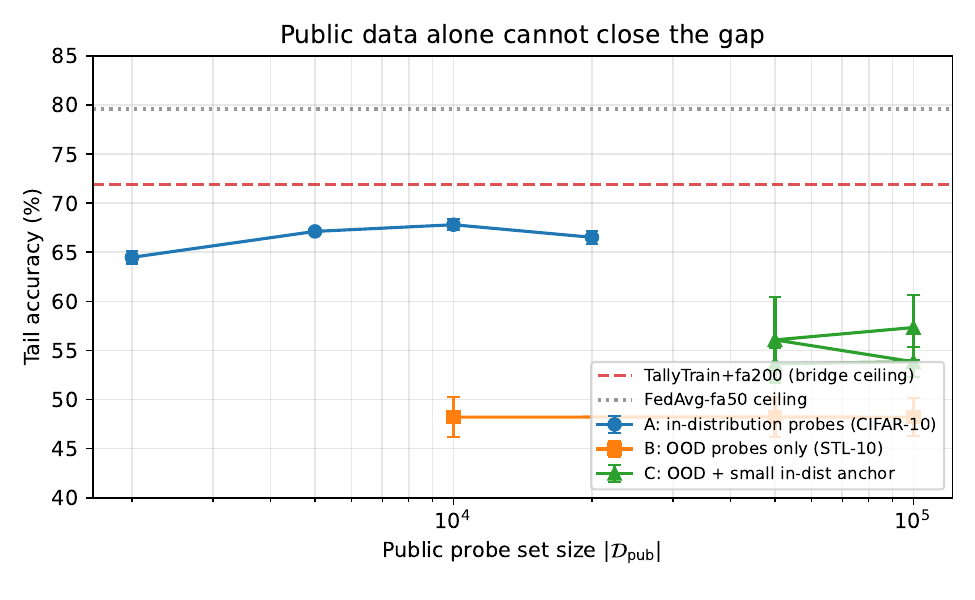}
\caption{Pure TallyTrain on CIFAR-10 non-IID with three families of
probe pools (3 seeds each). In-distribution
probes saturate around $67.5$\%; OOD probes alone near $48$\%; OOD
probes with a small in-distribution anchor only $\sim\!57$\%. The
bridge TallyTrain+fa$M$, even at $|\mathcal{D}_{\mathrm{pub}}|\!=\!2$k,
sits at $71.92$\% (red dashed) and FedAvg-fa50 at $79.6$\% (gray
dotted): probe abundance is not what closes the gap.}
\label{fig:dpub}
\end{figure}

Distillation between merges anchors peers in function space, so a
single merge can fully reconcile them. The bridge is not a workaround
for insufficient probe data: a separate sweep across in-distribution,
out-of-distribution and OOD$+$anchor probe pools
(Figure~\ref{fig:dpub}) shows that pure TallyTrain
saturates well below the bridge ceiling regardless of probe-pool
size. The parameter channel is the missing channel.

\subsection{Pareto frontier}
\label{sec:pareto}

\begin{figure}[!htbp]
\centering
\includegraphics[width=0.75\linewidth]{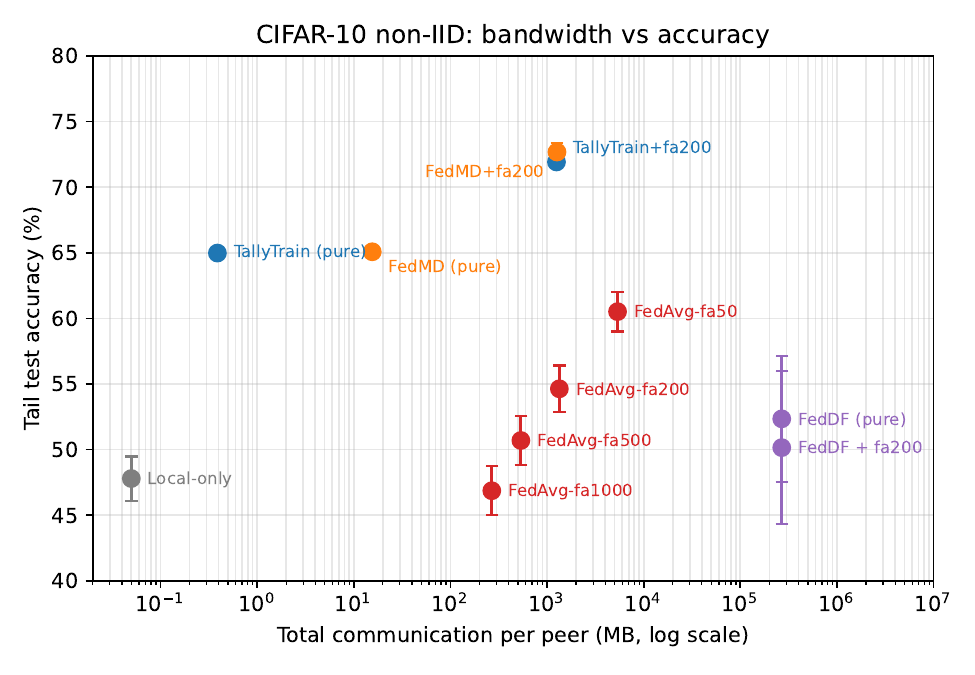}
\caption{Bandwidth-accuracy Pareto frontier, CIFAR-10 non-IID, 3
seeds (error bars omitted when smaller than the marker). Pure
TallyTrain anchors the low-bandwidth end; TallyTrain+fa$M$
dominates the upper-right.}
\label{fig:pareto}
\end{figure}

Figure~\ref{fig:pareto} plots tail accuracy vs.\ total per-peer
bandwidth (log scale). Pure TallyTrain anchors the low-bandwidth
end ($65.0\%$ at $0.39$~MB); TallyTrain+fa$M$ and FedMD+fa200 cluster
at the top-right and dominate every FedAvg/FedProx point at
$\sim\!1.3$~GB. FedProx is strictly dominated by FedAvg here
(Table~\ref{tab:t3}).

\subsection{Scaling with the number of output classes $C$}
\label{sec:scaling}

\begin{figure}[!htbp]
\centering
\includegraphics[width=0.75\linewidth]{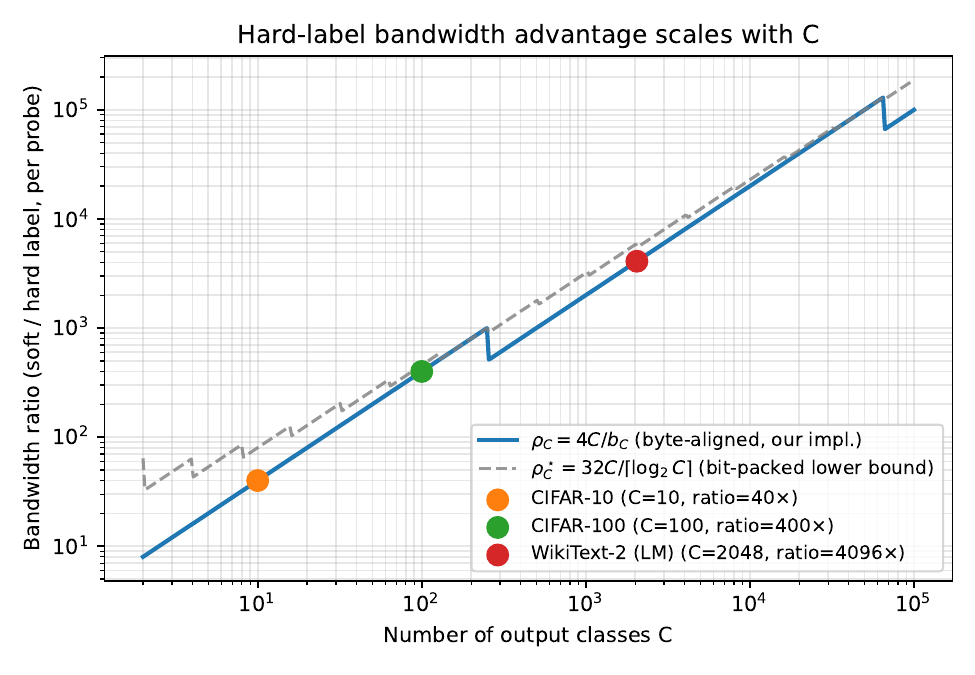}
\caption{Per-probe bandwidth ratio $\rho_C\!=\!4C/b_C$ between
32-bit soft labels and byte-aligned hard-label indices
($b_C\in\{1,2\}$), with empirical anchors at $C\!=\!10$ (CIFAR-10),
$C\!=\!100$ (CIFAR-100) and $C\!=\!2048$ (WikiText-2 BPE
vocabulary). The dashed curve is the bit-packed structural lower
bound $\rho_C^\star\!=\!32C/\lceil\log_2 C\rceil$.}
\label{fig:scaling}
\end{figure}

The byte-aligned per-probe bandwidth ratio is $\rho_C\!=\!4C/b_C$
with $b_C\!\in\!\{1,2\}$, giving $\rho_{10}\!=\!40$,
$\rho_{100}\!=\!400$, $\rho_{2048}\!=\!4096$
(Figure~\ref{fig:scaling}); the empirical Payload ratios in
Tables~\ref{tab:t1}~\&~\ref{tab:lm} match to within rounding. The bit-packed
lower bound $\rho_C^{\star}\!=\!32C/\lceil\log_2 C\rceil$ gives
$\rho_{2048}^{\star}\!\approx\!5957$.

\subsection{Secondary findings}
\label{sec:secondary}

\paragraph{Cross-distribution KL-decay.}
With CIFAR-100 probes for a CIFAR-10 task ($\alpha\!=\!0$, no useful
public labels), a linear KL-decay ($T\!=\!200$; robust across
$T\!\in\![50,400]$) reaches $48.35 \pm 2.10\%$ tail accuracy versus
FedMD's $49.45\%$ -- a statistical tie given the seed spread -- at
$40\times$ less bandwidth. Without decay, both
methods collapse to near-random as the consensus drifts into a
self-reinforcing fixed point.

\paragraph{Coverage rule and rank depth.}
The channel scales with $\kappa\!=\!B_{\mathrm{sample}}/C$:
$\kappa\!\in\![1,2]$ is optimal on CIFAR-10 and best-of-sweep on
CIFAR-100, $\kappa\!<\!0.5$
starves it (CIFAR-10 tail $65\!\to\!37\%$ at $\kappa\!=\!0.4$), and
$\kappa\!>\!5$ \emph{degrades} low-$C$ accuracy by over-weighting
public-set fit. Rank depth adds little: RC-FD with
$\mathrm{top}\text{-}k\!\in\!\{1,3,5\}$ gives $64.82, 65.20, 65.29\%$ on
CIFAR-10 hybrid, so 1 byte captures nearly all the softmax's
information in the labelled regime.

\subsection{Cross-modality validation: federated language modeling at $C = 2048$}
\label{sec:charlm}

To test whether the class-count-axis advantage transfers, we replicate the
headline cells on next-token prediction over WikiText-2
\cite{merity2016pointer}: $N\!=\!8$ peers, a tiny GPT (6 layers,
$d\!=\!256$, $\sim\!5.84$M parameters) trained from scratch on disjoint
chunks of the BPE-2048 corpus, with $5\%$ of train held out as the
public probe set. Each peer transmits one prediction per token
position; $B_{\mathrm{sample}}\!=\!32$ length-$256$ sequences gives
$8192$ probe positions per round. We report mean top-1 accuracy and perplexity on
the validation split. Among the language-modeling (LM) findings, the first three
reproduce our CIFAR results, the fourth is specific to this regime.

\begin{table}[t]
\centering
\small
\resizebox{0.72\textwidth}{!}{%
\begin{tabular}{l|cc|cc}
\toprule
Method & Tail acc.\ (\%) & Tail PPL & Payload & $\times$ vs.\ FedMD \\
\midrule
TallyTrain (pure) & $15.45 \pm 0.03$ & $10{,}087 \pm 45$ & \textbf{0.31 GB} & $\mathbf{4096\times}$ \\
FedMD (pure)         & $15.78 \pm 0.02$ & $\phantom{0}3{,}622 \pm 71$ & 1{,}268 GB & $1\times$ \\
\midrule
FedAvg-fa200         & $21.52 \pm 0.03$ & $\phantom{0,0}408.3 \pm 5.0$ & 0.70 GB & --- \\
\textbf{TallyTrain+fa200} & $22.97 \pm 0.03$ & $\phantom{0,0}403.3 \pm 6.0$ & \textbf{1.01 GB} & $\mathbf{1256\times}$ \\
FedMD+fa200          & $23.01 \pm 0.05$ & $\phantom{0,0}339.0 \pm 5.9$ & 1{,}269 GB & $1\times$ \\
\midrule
FedAvg-fa1           & $29.17 \pm 0.16$ & $\phantom{0,00}80.9 \pm 1.1$ & 140 GB & --- \\
\bottomrule
\end{tabular}%
}
\caption{WikiText-2 (BPE-2048), $N\!=\!8$, 3 seeds; tail metrics over
the last 600 rounds. Top: pure distillation; middle: bandwidth-bridge
(fa200) vs pure FedAvg. Hard-label TallyTrain matches soft-label FedMD
on accuracy in both regimes; the last column is the hard-vs-soft
bandwidth reduction (parameter-only FedAvg rows have no soft
counterpart). $\sigma\!\le\!0.16$~pp throughout. Interpretation in
LM-1--LM-4.}
\label{tab:lm}
\end{table}

\noindent\textbf{(LM-1) Hard $\approx$ soft at $C\!=\!2048$.} Pure
TallyTrain trails FedMD by $0.33$~pp -- $2\%$ relative -- at
$4096\times$ less bandwidth (Table~\ref{tab:lm}), confirming the
structural ratio $\rho_C\!=\!4C/b_C$ two orders of magnitude beyond
CIFAR-10.

\noindent\textbf{(LM-2) Both pure methods collapse on high-entropy
targets.} Accuracy peaks near round~450, then perplexity explodes by
one-to-two orders of magnitude -- the same Condorcet drift as the
cross-distribution image case (\S\ref{sec:secondary}), here
in-distribution because rare-token positions are intrinsically
high-entropy.

\noindent\textbf{(LM-3) The bandwidth-bridge variant transfers.}
TallyTrain$+$fa200 beats FedAvg-fa200 by $+1.45$~pp for $1.4\times$ the
bandwidth (Table~\ref{tab:lm});
pure TallyTrain does not catch the far costlier FedAvg-fa1, so the
Pareto-optimal low-bandwidth point is the bridge, not pure TallyTrain.

\noindent\textbf{(LM-4) Small model, large $C$: the channel choice
matters most.} The tiny GPT makes parameter merges cheap, so the
consensus channel is $31\%$ of the bridge budget here against $0.03\%$
on CIFAR-10, and the soft-label channel swamps FedMD's entirely. At matched accuracy,
FedMD$+$fa200 costs $1{,}256\times$ more than TallyTrain$+$fa200
(Table~\ref{tab:lm}) -- though FedMD wins perplexity, a soft-label
edge. Here the $\lceil\log_2 C\rceil$-bit index is exactly where the
saving comes from.

\section{Discussion}
\label{sec:discussion}

\subsection{The failure mode of pure-KL distillation, and its fix}
\label{sec:when_kl_fails}

\noindent\textbf{The failure mode is Condorcet-class collapse.}
Pure-KL distillation collapses whenever peer agreement on the public
probes is too weak for the contraction of Lemma~1 to reach the truth:
peers first agree on accurate predictions, then drift into a
self-reinforcing fixed point unrelated to the test distribution. This
same mechanism drives both collapses we report -- the
cross-distribution image case (\S\ref{sec:secondary}) and the
high-entropy WikiText-2 case (\S\ref{sec:charlm}, LM-2).
\textbf{The bridge variant fixes both without retuning:} a more
reliable anchor -- KL-decay for cross-distribution images, periodic
FedAvg merges for the LM (\S\ref{sec:t3}) -- halts the drift and yields
the TallyTrain+fa$M$ cells that dominate Tables~\ref{tab:t3}
and~\ref{tab:lm}.

\subsection{The bridge dominates parameter averaging}

\noindent\textbf{The bridge closes the absolute-accuracy gap.} Pure
TallyTrain plateaus $\sim\!14$~pp below the parameter-space ceiling on
both modalities. TallyTrain+fa$M$
closes that gap and dominates every FedAvg/FedProx/FedDF cell on
CIFAR-10, and every cell at comparable bandwidth on WikiText-2
(Tables~\ref{tab:t3},~\ref{tab:lm}) -- on WikiText-2 the win is
$+1.45$~pp for $1.4\times$ the bandwidth. This
lift is public-set supervision, not hard- over soft-label consensus
(\S\ref{sec:t3}); the bridge's role is to carry it over a
$\lceil\log_2 C\rceil$-bit channel that doubles as a between-merge
stabilizer.

\noindent\textbf{The bridge is also the most reproducible cell.}
Among bridge and parameter-averaging cells the bandwidth-bridge has the
smallest seed variance -- $\sigma\!=\!0.38$~pp on CIFAR-10 ($\ge\!1.52$
for any FedAvg, $\ge\!4.79$ for FedDF) -- and is essentially
deterministic on WikiText-2 ($\sigma\!=\!0.03$~pp): plurality voting is
invariant to logit noise, and merges further contract parameter spread.

\noindent\textbf{The bridge dominates the stronger baselines too, each
for a mechanistic reason.} FedProx (best $\mu\!=\!0.001$) sits
$1.7$--$2.9$~pp below plain FedAvg at every fa -- its proximal anchor
goes stale over the $\mathrm{fa}\!\cdot\!K$ inter-merge steps -- so the
bridge beats its best FedProx cell by $+13.1$~pp.
FedDF \citep{lin2020ensemble} peaks respectably ($66.0\!\pm\!1.6\%$)
but its tail collapses ($52.3\!\pm\!4.8$) at FedAvg-fa1's bandwidth --
its server-side distillation crashes on the same high-entropy targets,
and its central aggregator removes the voting filter -- so the bridge
dominates it by $19.6$~pp at $200\times$ less bandwidth.

\noindent\textbf{The hard-label channel turns FedAvg's
merge-frequency slack into a Pareto-dominant operating point.} Our \texttt{fa} merge cadence is exactly the
local-SGD averaging interval FedAvg exposes through its local-epoch
count $E$ (\S\ref{sec:algorithm}); accuracy degrades smoothly as it
grows (Table~\ref{tab:t3}), consistent with local-SGD bounds
\citep{stich2018local,khaled2020tighter}. The contribution is not that
low-frequency FedAvg works -- McMahan's own ablations and local-SGD
theory predict that -- but that a cheap $\lceil\log_2 C\rceil$-bit
consensus channel cleanly fills the inter-merge gap it leaves open, at
negligible bandwidth over the parameter merges themselves.

\subsection{Limitations and future work}
\label{sec:future}

\noindent\textbf{Scope of this evaluation.} Headline cells use
$N\!\le\!10$ homogeneous-architecture peers (matching FedMD's
$N\!=\!10$, FedDF's $N\!\le\!20$, and DiLoCo's $N\!\in\![8,32]$) under
full-mesh averaging on Dirichlet-$0.5$ non-IID splits -- the harder,
more practically relevant regime; the IID case should be at least as favourable but is
not measured. We do not validate $N\!\gg\!10$, sparse-gossip
topologies, or heterogeneous architectures.

\noindent\textbf{Four extensions.} (i) \emph{Sparse gossip}: the
full-mesh topology scales as $\Theta(N^2)$; sparse gossip would scale
as $\Theta(N\log N)$ and make decentralization strictly
bandwidth-favourable, though we have not tested the hard-label
primitive under restricted neighbourhoods. (ii) \emph{Adaptive
consensus weighting}: an online voting-margin estimator could replace
the fixed KL-decay of \S\ref{sec:secondary} and unify the labelled,
cross-distribution, and LM regimes. (iii) \emph{Heterogeneous
architectures}: the primitive is architecture-free by construction,
but the contraction lemma under model-class mismatch is open.
(iv) \emph{Production-scale LMs}: scaling our $C\!=\!2048$ tiny-GPT
anchor to $C\!=\!32{,}000$--$50{,}000$ vocabularies and
billion-parameter models would push $\rho_C$ past $60{,}000\times$; we
expect the TallyTrain+fa$M$ vs.\ FedAvg-fa$M$ ranking to persist, but
demonstrating it at scale is future work.

\section{Conclusion}

TallyTrain transmits only each peer's $\arg\max$ class index, beating
soft-label distillation on CIFAR-100 and matching it elsewhere at
$40\times$--$4{,}096\times$ less
bandwidth across CIFAR-10/100 and WikiText-2 and -- composed with
sparse FedAvg merges -- improving the accuracy--bandwidth trade-off of
every parameter-averaging point we tested. The accuracy gain comes from
the voted target staying peaked where the soft average flattens toward
uniform (Proposition~2; $+1.13$ vs $-0.22$~pp on CIFAR-100). The bandwidth gain
scales as $\rho_C\!=\!4C/b_C$, with no model-size term: a
billion-parameter model with the same vocabulary gets the same
per-probe reduction as our small ones, so the advantage tracks the
class count $C$, not the parameter count, and does not diminish with
scale; the per-probe channel is model-size-free, though the bridge's
parameter merges are not. It is especially attractive for
large-vocabulary settings current soft-label methods cannot reach.

\acks{We thank Dr.\ Mateusz Kozi\'nski for his valuable suggestions
that improved the paper.}

% jmlr.cls auto-sets bibliographystyle to plainnat; do not redeclare.
{\footnotesize
\setlength{\bibsep}{0pt plus 0.3ex}
\bibliography{references}

@inproceedings{hinton2015distilling,
  title     = {Distilling the Knowledge in a Neural Network},
  author    = {Hinton, Geoffrey and Vinyals, Oriol and Dean, Jeff},
  booktitle = {NIPS Deep Learning and Representation Learning Workshop},
  year      = {2015},
}

@article{fedmd2019,
  title = {FedMD: Heterogeneous Federated Learning via Model Distillation},
  author = {Li, Daliang and Wang, Junpu},
  journal = {arXiv preprint arXiv:1910.03581},
  year = {2019},
  note = {NeurIPS 2019 Workshop on Federated Learning for Data Privacy and Confidentiality},
}

@article{fedkd2021,
  title = {Communication-efficient Federated Learning via Knowledge Distillation},
  author = {Wu, Chuhan and Wu, Fangzhao and Lyu, Lingjuan and Huang, Yongfeng and Xie, Xing},
  journal = {Nature Communications},
  volume = {13},
  pages = {2032},
  year = {2022},
}

@article{dfml2024,
  title = {DFML: Decentralized Federated Mutual Learning},
  author = {Khalil, Yasser H. and Estiri, Amir Hossein and Beitollahi, Mahdi and Asadi, Nader and Hemati, Sobhan and Li, Xu and Zhang, Guojun and Chen, Xi},
  journal = {Transactions on Machine Learning Research},
  year = {2024},
}

@inproceedings{desa2024,
  title = {Overcoming Data and Model heterogeneities in Decentralized Federated Learning via Synthetic Anchors},
  author = {Huang, Chun-Yin and Srinivas, Kartik and Zhang, Xin and Li, Xiaoxiao},
  booktitle = {Proceedings of the 41st International Conference on Machine Learning},
  series = {Proceedings of Machine Learning Research},
  volume = {235},
  pages = {20111--20133},
  year = {2024},
  publisher = {PMLR},
}

@inproceedings{dspodfl2025,
  title = {Decentralized Sporadic Federated Learning: A Unified Algorithmic Framework with Convergence Guarantees},
  author = {Zehtabi, Shahryar and Han, Dong-Jun and Parasnis, Rohit and Hosseinalipour, Seyyedali and Brinton, Christopher},
  booktitle = {International Conference on Learning Representations},
  year = {2025},
  note = {Spotlight},
}

@inproceedings{dpfl2025,
  title = {DPFL: Decentralized Personalized Federated Learning},
  author = {Kharrat, Salma and Canini, Marco and Horv{\'a}th, Samuel},
  booktitle = {Proceedings of The 28th International Conference on Artificial Intelligence and Statistics},
  series = {Proceedings of Machine Learning Research},
  volume = {258},
  pages = {5086--5094},
  year = {2025},
  publisher = {PMLR},
}

@inproceedings{ntkdfl2025,
  title = {NTK-DFL: Enhancing Decentralized Federated Learning in Heterogeneous Settings via Neural Tangent Kernel},
  author = {Thompson, Gabriel and Yue, Kai and Wong, Chau-Wai and Dai, Huaiyu},
  booktitle = {Proceedings of the 42nd International Conference on Machine Learning},
  series = {Proceedings of Machine Learning Research},
  volume = {267},
  pages = {59470--59491},
  year = {2025},
  publisher = {PMLR},
}

@inproceedings{fedspd2025,
  title = {FedSPD: A Soft-clustering Approach for Personalized Decentralized Federated Learning},
  author = {Lin, I-Cheng and Yagan, Osman and Joe-Wong, Carlee},
  booktitle = {Proceedings of the Forty-first Conference on Uncertainty in Artificial Intelligence},
  series = {Proceedings of Machine Learning Research},
  volume = {286},
  pages = {2618--2641},
  year = {2025},
  publisher = {PMLR},
}

@inproceedings{fdpdfl2025,
  title = {Mitigating the Privacy--Utility Trade-off in Decentralized Federated Learning via $f$-Differential Privacy},
  author = {Li, Xiang and Su, Buxin and Wang, Chendi and Long, Qi and Su, Weijie},
  booktitle = {Advances in Neural Information Processing Systems 38},
  year = {2025},
}

@inproceedings{compadvdfl2025,
  title = {Competitive Advantage Attacks to Decentralized Federated Learning},
  author = {Jia, Yuqi and Fang, Minghong and Gong, Neil Zhenqiang},
  booktitle = {Advances in Neural Information Processing Systems 38},
  year = {2025},
}

@article{gflat2026,
  title = {Achieving Global Flatness in Decentralized Learning with Heterogeneous Data},
  author = {Choudhary, Sakshi and Aketi, Sai Aparna and Roy, Kaushik},
  journal = {Transactions on Machine Learning Research},
  year = {2026},
}

@article{pame2026,
  title = {Decentralized Federated Learning by Partial Message Exchange},
  author = {Sha, Shan and Zhou, Shenglong and Wang, Xin and Kong, Lingchen and Li, Geoffrey Ye},
  journal = {arXiv preprint arXiv:2603.01730},
  year = {2026},
}

@article{spodgt2026,
  title = {Sporadic Gradient Tracking over Directed Graphs: A Theoretical Perspective on Decentralized Federated Learning},
  author = {Zehtabi, Shahryar and Han, Dong-Jun and Hosseinalipour, Seyyedali and Brinton, Christopher},
  journal = {arXiv preprint arXiv:2602.00791},
  year = {2026},
}

@article{dfedcad2025,
  title = {On the Fast Adaptation of Delayed Clients in Decentralized Federated Learning: A Centroid-Aligned Distillation Approach},
  author = {Bai, Jiahui and Dong, Hai and Qin, A. K.},
  journal = {arXiv preprint arXiv:2508.02993},
  year = {2025},
}

@inproceedings{mcmahan2017communication,
  title = {Communication-Efficient Learning of Deep Networks from Decentralized Data},
  author = {McMahan, H. Brendan and Moore, Eider and Ramage, Daniel and Hampson, Seth and Ag{\"u}era y Arcas, Blaise},
  booktitle = {Proceedings of the 20th International Conference on Artificial Intelligence and Statistics (AISTATS)},
  series = {Proceedings of Machine Learning Research},
  volume = {54},
  pages = {1273--1282},
  year = {2017},
}

@inproceedings{douillard2023diloco,
  title     = {{DiLoCo}: Distributed Low-Communication Training of Language Models},
  author    = {Douillard, Arthur and Feng, Qixuan and Rusu, Andrei Alex and Chhaparia, Rachita and Donchev, Yani and Kuncoro, Adhiguna and Ranzato, MarcAurelio and Szlam, Arthur and Shen, Jiajun},
  booktitle = {ICML 2024 Workshops: WANT},
  year      = {2024},
}

@inproceedings{stich2018local,
  title = {Local {SGD} Converges Fast and Communicates Little},
  author = {Stich, Sebastian U.},
  booktitle = {International Conference on Learning Representations},
  year = {2019},
}

@inproceedings{khaled2020tighter,
  title = {Tighter Theory for Local {SGD} on Identical and Heterogeneous Data},
  author = {Khaled, Ahmed and Mishchenko, Konstantin and Richt{\'a}rik, Peter},
  booktitle = {Proceedings of the 23rd International Conference on Artificial Intelligence and Statistics (AISTATS)},
  series = {Proceedings of Machine Learning Research},
  volume = {108},
  pages = {4519--4529},
  year = {2020},
}

@inproceedings{orbitdfl2026,
  title = {Learning in Orbit: A Physics-Aware Graph-Decentralized Federated Learning for Multi-Satellite Space Situational Awareness},
  author = {Kaur, Gagandeep and Prasad, Ranjitha},
  booktitle = {AAAI 2026 Workshop on Federated Learning and Collaborative AI (FLCA)},
  year = {2026},
}

@inproceedings{lin2020ensemble,
  title     = {Ensemble Distillation for Robust Model Fusion in Federated Learning},
  author    = {Lin, Tao and Kong, Lingjing and Stich, Sebastian U. and Jaggi, Martin},
  booktitle = {Advances in Neural Information Processing Systems},
  volume    = {33},
  pages     = {2351--2363},
  year      = {2020},
}

@article{kairouz2019advances,
  title   = {Advances and Open Problems in Federated Learning},
  author  = {Kairouz, Peter and McMahan, H. Brendan and Avent, Brendan and Bellet, Aur{\'e}lien and Bennis, Mehdi and others},
  journal = {Foundations and Trends in Machine Learning},
  volume  = {14},
  number  = {1--2},
  pages   = {1--210},
  year    = {2021},
}

@inproceedings{li2018federated,
  title     = {Federated Optimization in Heterogeneous Networks},
  author    = {Li, Tian and Sahu, Anit Kumar and Zaheer, Manzil and Sanjabi, Maziar and Talwalkar, Ameet and Smith, Virginia},
  booktitle = {Proceedings of Machine Learning and Systems},
  volume    = {2},
  year      = {2020},
}

@article{jeong2018communication,
  title={Communication-Efficient On-Device Machine Learning: Federated Distillation and Augmentation under Non-IID Private Data},
  author={Jeong, Eunjeong and Oh, Seungeun and Kim, Hyesung and Park, Jihong and Bennis, Mehdi and Kim, Seong-Lyun},
  journal={arXiv preprint arXiv:1811.11479},
  year={2018}
}

@inproceedings{chang2019cronus,
  title     = {Cronus: Robust and Heterogeneous Collaborative Learning with Black-Box Knowledge Transfer},
  author    = {Chang, Hongyan and Shejwalkar, Virat and Shokri, Reza and Houmansadr, Amir},
  booktitle = {1st NeurIPS Workshop on New Frontiers in Federated Learning},
  year      = {2021},
}

@inproceedings{merity2016pointer,
  title     = {Pointer Sentinel Mixture Models},
  author    = {Merity, Stephen and Xiong, Caiming and Bradbury, James and Socher, Richard},
  booktitle = {International Conference on Learning Representations},
  year      = {2017},
}

@inproceedings{korda2016distributed,
  title     = {Distributed Clustering of Linear Bandits in Peer to Peer Networks},
  author    = {Korda, Nathan and Sz{\"o}r{\'e}nyi, Bal{\'a}zs and Li, Shuai},
  booktitle = {Proceedings of the 33rd International Conference on Machine Learning (ICML)},
  series    = {Proceedings of Machine Learning Research},
  volume    = {48},
  pages     = {1301--1309},
  year      = {2016},
}

@inproceedings{mahadik2020fast,
  title     = {Fast Distributed Bandits for Online Recommendation Systems},
  author    = {Mahadik, Kanak and Wu, Qingyun and Li, Shuai and Sabne, Amit},
  booktitle = {Proceedings of the 34th ACM International Conference on Supercomputing (ICS)},
  year      = {2020},
}

@inproceedings{narasimhan2016stochastic,
  title     = {Stochastic Optimization Techniques for Quantification Performance Measures},
  author    = {Narasimhan, Harikrishna and Li, Shuai and Kar, Purushottam and Chawla, Sanjay and Sebastiani, Fabrizio},
  booktitle = {Proceedings of the 22nd ACM SIGKDD International Conference on Knowledge Discovery and Data Mining (KDD)},
  year      = {2016},
}
}

\end{document}